\documentclass{article} % For LaTeX2e
\usepackage[nonatbib,preprint]{neurips_2020}

\usepackage{amsmath,amsfonts,bm}

\def\eqref#1{equation~\ref{#1}}
\def\1{\bm{1}}

\DeclareMathAlphabet{\mathsfit}{\encodingdefault}{\sfdefault}{m}{sl}
\SetMathAlphabet{\mathsfit}{bold}{\encodingdefault}{\sfdefault}{bx}{n}

\DeclareMathOperator*{\argmax}{arg\,max}

\usepackage{hyperref}
\usepackage{url}
\usepackage{algorithm}
\usepackage[noend]{algpseudocode}
\usepackage{graphicx}
\usepackage{multirow}
\usepackage{wrapfig}
\usepackage[T1]{fontenc}
\usepackage{mathtools}
\usepackage{subcaption}
\usepackage{amsthm,amssymb, amsmath}
\usepackage[utf8]{inputenc} % allow utf-8 input
\usepackage[T1]{fontenc}    % use 8-bit T1 fonts
\usepackage{hyperref}       % hyperlinks
\usepackage{url}            % simple URL typesetting
\usepackage{booktabs}       % professional-quality tables
\usepackage{amsfonts}       % blackboard math symbols
\usepackage{nicefrac}       % compact symbols for 1/2, etc.
\usepackage{microtype}      % microtypography
\usepackage{color}          % color
\usepackage{enumitem}
\usepackage[numbers]{natbib}
\usepackage{bbm}
\newcommand{\algalign}[2]% #1 = text to left, #2 = text to right
{\makebox[\maxwidth][r]{$#1{}$}${}#2$}
\title{Gaussian MRF Covariance Modeling for Efficient Black-Box Adversarial Attacks}
\author{Anit Kumar Sahu\\
Bosch Center for Artificial Intelligence\\
\texttt{anit.sahu@gmail.com} \\
\And
Satya Narayan Shukla \\
College of Information and Computer Sciences\\
University of Massachusetts Amherst\\
\texttt{snshukla@cs.umass.edu} \\
\And
J. Zico Kolter \\
Bosch Center for Artificial Intelligence\\
Carnegie Mellon University\\
\texttt{zkolter@cs.cmu.edu} \\
}
\begin{document}

\maketitle

\begin{abstract}
We study the problem of generating adversarial examples in a black-box setting, where we only have access to a zeroth order oracle, providing us with loss function evaluations. Although this setting has been investigated in previous work, most past approaches using zeroth order optimization implicitly assume that the gradients of the loss function with respect to the input images are \emph{unstructured}. In this work, we show that in fact substantial correlations exist within these gradients, and we propose to capture these correlations via a Gaussian Markov random field (GMRF). Given the intractability of the explicit covariance structure of the MRF, we show that the covariance structure can be efficiently represented using the Fast Fourier Transform (FFT), along with low-rank updates to perform exact posterior estimation under this model.  We use this modeling technique to find fast one-step adversarial attacks, akin to a black-box version of the Fast Gradient Sign Method~(FGSM), and show that the method uses fewer queries and achieves higher attack success rates than the current state of the art.  We also highlight the general applicability of this gradient modeling setup.

% We employ Markov Random Fields~(MRFs) to exploit the structure of input data to systematically model the covariance structure of the gradients. The MRF structure in addition to Bayesian inference for the gradients facilitates one-step attacks akin to Fast Gradient Sign Method~(FGSM) albeit in the black-box setting. The resulting method uses fewer queries than the current state of the art to achieve comparable performance. In particular, in the regime of lower query budgets, we show that our method is particularly effective in terms of fewer average queries with high attack accuracy while employing one-step attacks.
\end{abstract}

\section{Introduction}

Most methods for adversarial attacks on deep learning models operate in the so-called \emph{white-box} setting~\citep{goodfellow2014explaining}, where the model being attacked, and its gradients, are assumed to be fully known.  Recently, however there has also been considerable attention given to the \emph{black-box} setting as well, where the model is unknown and can only be queried by a user, and which much better captures the ``typical'' state by which an attacker can interact with a model~\citep{zoo,autozoom,bandits,parsimonious}. And several past methods in this area have conclusively demonstrated that, given sufficient number of queries, it is possible to achieve similarly effective attacks in the black-box setting akin to the white-box setting.  However, as has also been demonstrated by past work~\citep{bandits,nes,andriushchenko2019square}, the efficiency of these black-box attacks (the number of queries need to find an adversarial example) is fundamentally limited unless they can exploit the spatial correlation structure inherent in the model's gradients.  Yet, at the same time, most previous methods have used rather ad-hoc methods of modeling such correlation structure, such as using ``tiling'' bases and priors over time~\citep{bandits} or explicit square regions of the input space \citep{andriushchenko2019square} that require attack vectors be constant over large regions, or by other means such as using smoothly-varying perturbations~\citep{nes} to estimate these gradients.

In this work, we present a new, more principled approach to model the correlation structure of the gradients within the black-box adversarial setting.  In particular, we propose to model the gradient of the model loss function with respect to the input image using a Gaussian Markov Random Field (GMRF). This approach offers a number of advantages over prior methods: 1) it naturally captures the spatial correlation observed empirically in most deep learning models; 2) using the model, we are able to compute exact posterior estimates of the true gradient given observed data, while also fitting the parameters of the GMRF itself via an expectation maximization~(EM) approach; and 3) the method provides a natural alternative to uniformly sampling perturbations, based upon the eigenvectors of the covariance matrix.  Although representing the joint covariance over the entire input image may seem intractable for large-scale images, we can efficiently compute necessary terms for very general forms of grid-based GMRFs using the Fast Fourier Transform (FFT).

We evaluate our approach by attempting to find adversarial examples, over multiple different data sets and model architectures, using the GMRF combined with a very simple greedy zeroth order search technique; the method effectively forms a ``black-box'' version of the fast gradient sign method (FGSM), by constructing an estimate of the gradient at the input image itself, then taking a single signed step in this direction. Despite its simplicity, we show that owing to the correlation structure provided by the GMRF model, the approach outperforms more complex gradient-based approaches such as the \textsc{Bandits-td} \citep{bandits} or \textsc{Parsimonious} \citep{parsimonious} methods, especially for small query budgets.

\section{Related Work}
Most black-box adversarial attack methods catered towards untargeted attacks either rely on multi iteration optimization schemes or transfer-style attacks using substitute models.
Black-box adversarial attacks can be broadly categorized across a few different dimensions: optimization-based versus transfer-based attacks, and score-based versus decision-based attacks.

In the optimization-based adversarial setting, the adversarial attack is formulated as the problem of maximizing some loss function (e.g., the accuracy of the classifier or some continuous variant) using a zeroth order oracle, i.e., by making queries to the classifier.  And within this optimization setting, there is an important distinction between score-based attacks, which directly observe a traditional model loss, class probability, or other continuous output of the classifier on a given example, versus decision-based attacks, which only observe the hard label predicted by the classifier.  Decision based attacks have been studied by much past research \citep{brendel2017decision,Chen2019HopSkipJumpAttackAQ,zoo,shukla2019black}, and (not surprisingly) typically require more queries to the classifier than the score-based setting.  
Most black-box adversarial attacks can be categorized into score based and decision based attacks.  Multi iteration optimization based methods typically involve algorithms which access a zeroth order oracle, i.e., make queries to the model to obtain loss function evaluations in order to maximize the loss function so as to cause misclassification. Decision-based attacks, where the attacker only has access to the predicted label of the input by the model has been studied in \citet{brendel2017decision,Chen2019HopSkipJumpAttackAQ}.

In the setting of score-based attacks, the first such iterative attack on a class of binary classifiers was first studied in~\citep{nelson2012query}. A real-world application of black-box attacks to fool a PDF malware classifier was demonstrated by \citep{xu2016automatically}, for which a genetic algorithm was used. \citet{narodytska2017simple} demonstrated the first black-box attack on deep neural networks. Subsequently black-box attacks based on zeroth order optimization schemes, using techniques such as KWSA~\citep{kiefer1952stochastic} and RDSA~\citep{nesterov2017random} were developed in \citep{zoo,nes}. Though \citet{zoo} generated successful attacks attaining high attack accuracy, the method was found to be extremely query hungry which was then remedied to an extent by \citet{nes}. In \citep{bandits}, the authors exploit correlation of gradients across iterations by setting a prior and use a piece wise constant perturbation, i.e., tiling to develop a query efficient black-box method. In \citep{al2019there}, the authors focused on  estimating the signed version of the gradient to generate efficient black-box attacks. Recently \citet{parsimonious} used a combinatorial optimization perspective to address the black-box adversarial attack problem.

A concurrent line of work \citep{papernot2017practical} has considered the transfer-based setting. These approaches create adversarial attacks by training a substitute network with the aim to mimic the target model's decisions, which are then obtained through black-box queries. However, substitute network based attack strategies have been found to have a higher query complexity than those based on gradient estimation.

The exploitation of the structure of the input data space so as to append a regularizer has been recently found to be effective for robust learning. In particular, in \citep{lin2019wasserstein} showed that by using Wasserstein-2 geometry to capture semantically meaningful neighborhoods in the space of images helps to learn discriminative models that are robust to in-class variations of the input data.

\paragraph{Setting of this work} In this paper, we are specifically focused on the optimization-based, score-based setting, following most directly upon the work of \citet{zoo,nes,bandits,parsimonious}.  However, our contribution is also largely orthogonal to the methods presented in these prior works, and indeed could be combined with any gradient-based method.  Specifically, we show that by modeling the covariance structure of the gradient using a Gaussian MRF, a very simple approach achieves performance that is competitive with the best current methods in low query budgets. We further emphasize that this GMRF approach can be applied to other black-box search strategies as well.

\section{Preliminaries: Adversarial Attacks}
In the context of classifiers, adversarial examples are carefully crafted inputs to the classifier which have been perturbed by an additive perturbation so as to cause the classifier to misclassify the input. Formally, we define a classifier $C:\mathcal{X}\mapsto \mathcal{Y}$ and its corresponding classification loss function $L(\mathbf{x},y)$ (typically taken to be the cross-entropy loss on the class logits produced by the classifier), where $\mathbf{x}\in\mathcal{X}$ is the input to the classifier, $y\in\mathcal{Y}$, $\mathcal{X}$ is the set of inputs and $\mathcal{Y}$ is the set of labels. The objective of generating a misclassified example can be posed as an optimization problem. In particular, the aim is to generate an adversarial example $\mathbf{x}'$ for a given input $\mathbf{x}$ which maximizes $L(\mathbf{x}',y)$ but still remains $\epsilon_p$-close in terms of a specified metric to the original input. Thus, the generation of an adversarial attack can be formalized as a constrained optimization as follows:
\begin{align}
\label{eq:opt}
\mathbf{x}' = \argmax_{\mathbf{x}' :  \|\mathbf{x}' - \mathbf{x}\|_p \leq \epsilon_p} L(\mathbf{x}',y).
\end{align}
We give a brief overview of adversarial attacks categorized in terms of access to information namely, white-box and black-box attacks.

\paragraph{White-box attacks.} White-box settings assume access to the entire classifier and the analytical form of the (possibly non-convex) classifier's loss function. One of the original methods for generating such attacks is the Fast Gradient Sign Method~(FGSM) \citep{goodfellow2014explaining}, which remains relatively successful at attacking undefended models.  FGSM forms the attack using a single steepest descent update under the $\ell_\infty$ norm (similar analogues exist for other norms, though are technically then no longer gradient ``sign'' method), which corresponds to the update
\begin{align}
    \label{eq:fgsm}
     \mathbf{x}'=\mathbf{x}+\epsilon_{p}\textrm{sign}\left(\nabla L(\mathbf{x},y)\right).
\end{align}

A stronger and slightly more general attack uses the PGD algorithm to generate adversarial examples \citep{madry2018towards} (which was first introduced in the adversarial examples setting as the Basic Iterative method \citep{kurakin2016adversarial} in the $\ell_\infty$ setting), which corresponds to repeating the the updates
\begin{align}
  \label{eq:pgd_wb} 
  \mathbf{x}_{l} = \Pi_{B_p(\mathbf{x}, \epsilon)} (\mathbf{x}_{l-1} + \eta \mathbf{s}_l) ~~\textrm{with}~~
  \mathbf{s}_{l} = \Pi_{\partial B_p(0, 1)} \nabla_{x} L(\mathbf{x}_{l-1}, y),
\end{align}
where $\Pi_{S}$ denotes the projection onto the set $S$, $B_p(\mathbf{x}', \varepsilon')$ is
the $\ell_p$ ball of radius $\varepsilon'$ centered at $\mathbf{x}'$, $\eta$ denotes the step size,
and $\partial U$ is the boundary of a set $U$ (projection onto this set, e.g. takes the sign of the gradient for the $\ell_\infty$ ball, or normalizes the graident for the $\ell_p$ ball). 
By making, $\mathbf{s}_l$ to be the projection of the gradient
$\nabla_x L(\mathbf{x}_{l-1}, y)$ at $x_{l-1}$ onto the unit $\ell_p$ ball, it is ensured that $\mathbf{s}_{l}$ is the unit $\ell_p$-norm vector that has the
largest inner product with $\nabla_x L(\mathbf{x}_{l-1}, y)$.
However, in most real world deployments, it is impractical to assume complete access to the classifier and analytic form of the corresponding loss function, which makes black-box settings more realistic.

\paragraph{Black-box atacks.} In a typical {\em black-box} setting, the adversary's access is limited to the value of the loss function $L(\mathbf{x},y)$ for an input $(\mathbf{x},y)$. The crux of black-box methods can be divided into two classes. The first class consists of methods which operate in a derivative free manner, i.e., directly seek for the next iterate without estimating gradients. Evolution strategies such as CMA \citep{hansen2001completely} or alternatively Bayesian optimization method  \citep{shahriari2015taking} could be used for such attacks.  The second class more directly attempts estimate the gradient through, where our method is situated.

The main building block of gradient-based black-box methods is the use of some difference schemes to estimate gradients. A typical two-sided finite difference scheme, called the random direction stochastic approximation (RDSA), for instance, computes the estimate
\begin{align}
  \label{eq:rdsa} \widehat{\nabla}_x L(\mathbf{x}, y) = \frac{1}{m}\sum_{k=1}^m \mathbf{z}_k \left(L(\mathbf{x}+\delta \mathbf{z}_k, y)-L(\mathbf{x}, y)\right)/\delta,
\end{align}
where directions $\{\mathbf{z}_k\}_{k=1}^m$ can be canonical basis vectors, or normally distributed vectors, or the Fast Fourier Transform basis~(FFT) basis. The step size \mbox{$\delta > 0$} is a key parameter of choice; a higher $\delta$ could lead to extremely biased estimates, while a lower $\delta$ can lead to an unstable estimator. 
Owing to the biased gradient estimates, a finite difference based black-box attack turns out to be query hungry. 
In particular, in order to ensure sufficient increase of the objective at each iteration, the query complexity scales with dimension and hence is prohibitively large.

In this setting, the crux of our approach we present next is a method for \emph{decreasing} the variance of these gradient estimates, by exploiting the spatial correlation that arises due to the structure of deep networks. We then apply our approach within the context of a simple FGSM-like method, illustrating that even a simple algorithm, when combined with these more efficient gradient methods, can outperform many existing approaches in the low-query setting. In principle, however, our gradient estimation method can be combined with any adversarial attack that attempts to estimate gradients.

\section{Gradient Correlation and Query Efficient Black-Box Attacks}
\label{sec:prop}
In this section, we present the main algorithmic contribution of our work, a method for efficiently exploiting correlation to estimate network gradients, plus an application to a single-step FGSM-like attack.

\subsection{Gradient Correlation}
\begin{figure}[t]
\centering
\includegraphics[scale=0.3]{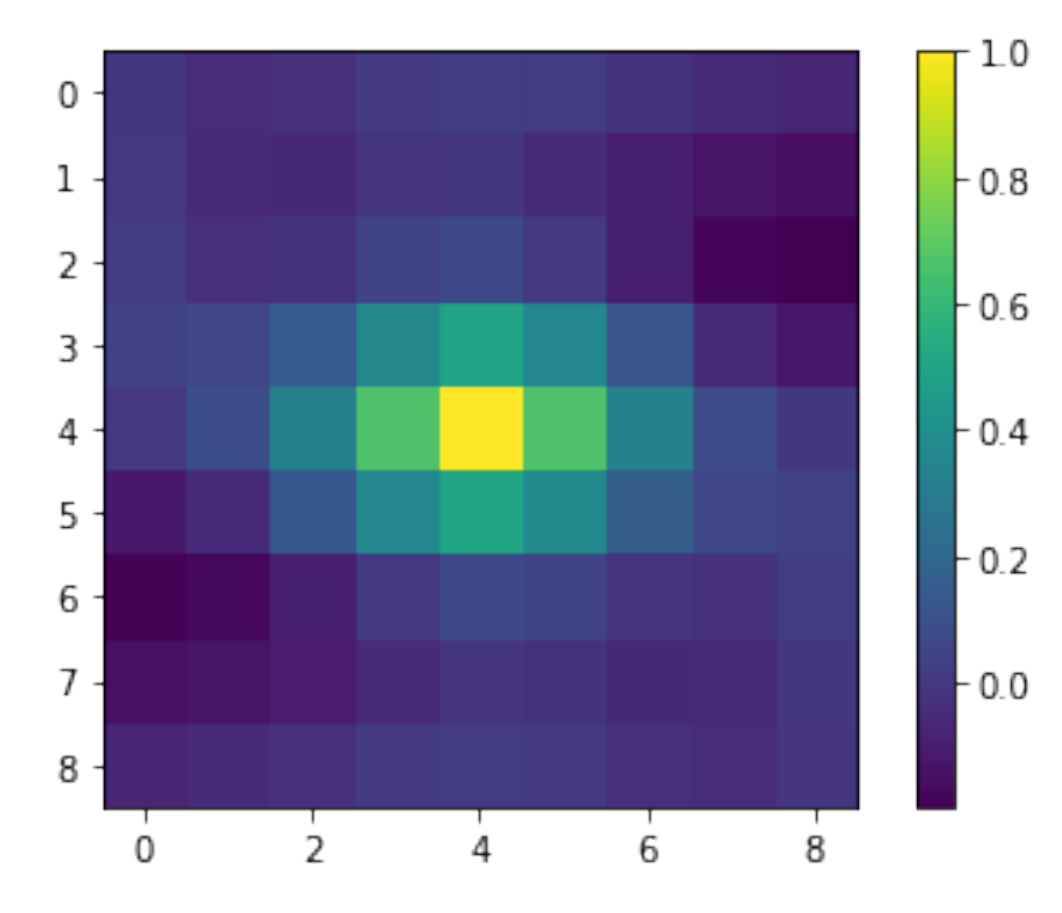}
\includegraphics[scale=0.3]{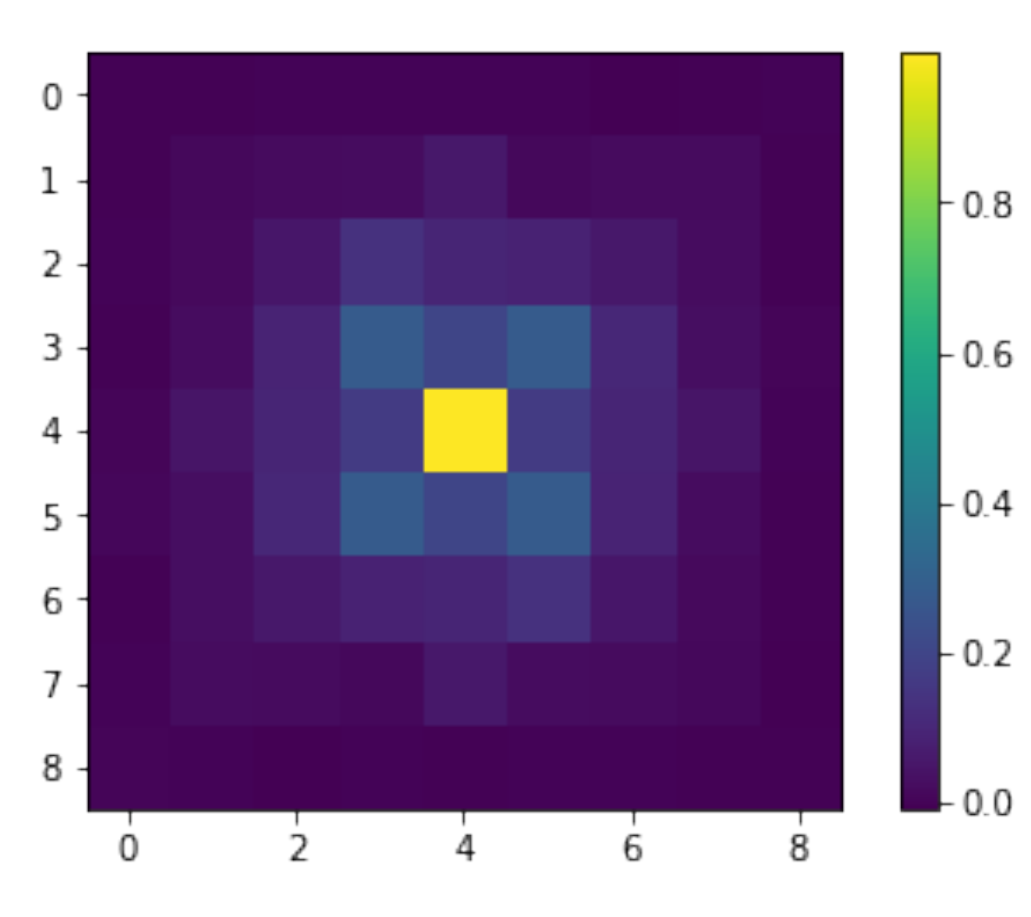}
\caption{Gradient autocorrelation shown for a 9x9 spatial window for (left) 100 MNIST images with a LeNet-like architecture; and (right) 50 ImageNet images on the VGG16 model for one of the color channels.}
\label{fig:autocorr}
\end{figure}

In most black-box adversarial attacks, the gradient terms across different images are implicitly assumed to be independent from each other. However, empirically we find that the gradients across different images exhibit substantial correlation across dimensions and across images.
In Figure \ref{fig:autocorr}, we plot the autocorrelation over a $9 \times 9$ window for the true gradients from $100$ images sampled from the MNIST test set on a LeNet model, for $50$ images sampled from the ImageNet validation set for the VGG16-bn model.  In both cases it is evident that there exists substantial correlation between nearby points.

We propose to exploit and model these correlations using a Gaussian Markov random field so as to reduce the variance of our gradient estimate.
Formally, letting $\mathbf{x}$ be the input to a classifier, and $\mathbf{g} = \nabla L(\mathbf{x},y)$ the gradient of the loss function with respect to the input, we are attempting to query loss function values and estimate the gradient, then we aim to put a prior distribution over $\mathbf{g}$
\begin{align}
\mathbf{g}\sim \mathcal{N}(0,\Sigma)
\end{align}
where $\Sigma$ is a non-identity covariance matrix modeling the correlation between terms. Following common practice we model the inverse covariance matrix $\Lambda = \Sigma^{-1}$, a setting also known as the Gaussian Markov random field~(GMRF) setting, given that the non-zero entries in $\Lambda$ correspond exactly to the edges in a graphical model describing the distribution.  Furthermore, we use a parameterized Gaussian MRF with relatively few free parameters.  For example, if $\mathbf{x}$ is a 2D image, then we may have one parameter $\alpha$ governing the diagonal terms $\Lambda_{i,i} = \alpha, \forall i$, and another governing adjacent pixels $\Lambda_{i,j} = \beta$ for $i,j$ corresponding to indices that are neighboring in the original image.  We will jointly refer to all the parameters of this model as $\theta$, so in this case $\theta = (\alpha,\beta)$, and we refer to the resulting $\Lambda$ as $\Lambda(\theta)$.

\subsection{Learning Gradient Correlation}
We first consider the problem of fitting a parameterized MRF model to given inputs $\mathbf{x}^{(1)},\ldots,\mathbf{x}^{(m)}$.  We further apply $n$ random directions $\mathbf{u}^{(1)}, \ldots, u^{(n)}$ to each image to produce multiple RDSA-like gradient estimates for each image $\mathbf{G}=[\mathbf{g}^{(1)},\ldots,\mathbf{g}^{(mn)}]$
where
\begin{equation}
    \mathbf{g}^{(ij)} = \mathbf{u}^{(j)} (L(\mathbf{x}^{(i)}+\delta \mathbf{u}^{(j)}, y^{(i)}) - L(\mathbf{x}^{(i)},y^{(i)}))/\delta.
\end{equation}
Given this data, the maximum likelihood estimation for the MRF parameters is simply the standard Gaussian maximum likelihood estimate, given by the optimization problem
\begin{align}
\label{eq:gmrf_obj}
\min_{\theta} \; \mathrm{tr} (\mathbf{S} \Lambda(\theta)) - \mathrm{logdet} (\Lambda(\theta)),
\end{align}
where $\mathbf{S} = \frac{1}{mn}\sum_{i} \mathbf{g}^{(i)} {\mathbf{g}^{(i)}}^{\top}$ is the sample covariance and $\mathrm{logdet}$ denotes the log determinant.  While this is a trivial problem for the case of general covariance, when we use a parameterized form of $\Lambda$, it becomes less clear how to solve this optimization problem efficiently for large inputs.

As we show, however, this optimization problem can be easily solved using the Fourier Transform; we focus for simplicity of presentation on the 2D case, but the method is easily generalizable to three dimensional convolutions to capture color channels in addition to the spatial dimensions itself. First, we focus on evaluating the trace term. The key idea here is that the $\Lambda$ operator can be viewed as a (circular) convolution\footnote{The FFT operation technically operates circular convolutions (meaning the convolution wraps around the image), and thus the covariance naively models a correlation between, e.g., the first and last rows of an image.  However, this is a minor issue in practice since: 1) it can be largely mitigated by zero-padding the input image before applying the FFT-based convolution, and 2) even if ignored entirely, the effect of a few additional circular terms in the covariance estimation is minimal.}
\begin{align}
\mathbf{K}(\theta) = \left [ \begin{array}{ccc}0 & \beta & 0 \\ \beta & \alpha & \beta \\ 0 & \beta & 0 \end{array} \right ],
\end{align}
which then lets us compute $\mathrm{tr} (S \Lambda(\theta))$ as sum of the elements of the product of $\mathbf{G}$ and the zero padded 2D convolution of $\mathbf{G}$ and $\mathbf{K}(\theta)$. For the log determinant term, we again use the fact that $\Lambda$ is a convolution operator.  Specifically, because it is a convolution, it can be diagonalized using the fast Fourier transform
\begin{align*}
\Lambda(\theta) = \mathbf{Q}^H \mathbf{D} \mathbf{Q}
\end{align*}
where $\mathbf{Q}$ is the the Fast Fourier Transform (FFT) basis, and the eigenvalues being the diagonal elements of $\mathbf{D}$ can be found by a FFT to the zero-padded convolution operator; thus, we can compute the log determinant term by simply taking the sum of the log of the FFT-computed eigenvalues. Once the efficient computation of the objective function is in place, we then employ Newton's method to optimize the objective, with the gradient and Hessian terms all computed via automatic differentiation (since there typically very few parameters, even computing the full Hessian matrix is straightforward). The entire procedure of estimating the GMRF parameters is depicted in Algorithm \ref{alg:obj}. Other appropriate optimization schemes could also be used instead of Newton's method, if desired.

For a $N \times N$ sized image, the dominating cost for the procedure is the $O(N^2 \log N)$ computation of the 2D FFT; this contrasts with the $O(N^6)$ naive complexity of the eigen decomposition of the $N^2 \times N^2$ inverse covariance.  Thus, the use of the properties of convolution operator and FFT basis makes the optimization in \eqref{eq:gmrf_obj} to be feasible: for example, computing the log determinant term for the VGG16-bn architecture for ImageNet would involve performing eigenvalue decomposition of a matrix with dimensions $150528\times150528$, which is computationally prohibitive without using the FFT basis.
\begin{algorithm}[t]
\caption{Learning GMRF Parameters}
\label{alg:obj}
\begin{algorithmic}[1]
\Procedure{\textsc{LearnGMRF}}{$\{\mathbf{x}^{(i)},y^{(i)}\}_{i=1}^{m},\delta$}
  %\settowidth{\maxwidth}{$m$}% use the widest one
  \State Draw $n$ vectors $\mathbf{u}^{(1)},\ldots,\mathbf{u}^{(n)}\sim\mathcal{N}(0,\mathbf{I})$
  \State Compute gradient estimates $\mathbf{g}^{(1)},\ldots,\mathbf{g}^{(mn)}$ where \vspace{-0.1in}\[
  \mathbf{g}^{(ij)} = \mathbf{u}^{(j)} (L(\mathbf{x}^{(i)}+\delta \mathbf{u}^{(j)}, y^{(i)}) - L(\mathbf{x}^{(i)},y^{(i)}))/\delta\]
  
  %\State Generate $\widehat{\mathbf{G}}$ from $\mathbf{G}$ by concatenating along each dimension.
  \While{Not converged}
  \State Compute objective: \vspace{-0.1in}\[
  \begin{split}
  &f(\theta)  = \mathrm{tr} (\mathbf{S} \Lambda(\theta)) - \mathrm{logdet} (\Lambda(\theta)) \nonumber\\&= \textrm{sum}\left(\mathbf{G}\times \textrm{conv2d}(\mathbf{K}(\theta),\mathbf{G})\right)+ \textrm{sum} \left (\log(\textrm{FFT}(\textrm{Pad}(\mathbf{K}(\theta)))) \right) \end{split} \]
  \State Perform Newton update: $\theta := \theta - \nabla^2 f(\theta)^{-1} \nabla f(\theta)$ (using automatic differentiation)
  \EndWhile
  \State \Return $\theta$
\EndProcedure
\end{algorithmic}
\end{algorithm}

\subsection{Application to Estimating Adversarial Directions}
\label{subsec:adv_est}
With the GMRF estimation framework in place, our next task is to estimate adversarial directions which involves estimating the posterior mean. 
Next, we apply the GMRF-based gradient estimate to the setting of constructing black-box attacks in a query efficient manner.  As we show later, exploiting the correlation across gradients substantially reduces query complexity without compromising on the attack success rate.  Under the aforementioned GMRF framework, we can interpret black-box gradient estimation as a Gaussian inference problem.
Specifically, we have assumed that the gradient at an input $\mathbf{x}$ follows the normal distribution with the prescribed inverse covariance
\begin{align*}
\mathbf{g} \sim \mathcal{N}(0,\Lambda^{-1}).
\end{align*}
We next construct a linear observation model by observing the fact that the loss function value at some point $\mathbf{x}'$ can be viewed as a noisy observation of a linear function of the gradient
\begin{align*}
L(\mathbf{x}') \approx L(\mathbf{x}) + \mathbf{g}^{\top} (\mathbf{x}' - \mathbf{x}),
\end{align*}
where we drop the $y$ in $L(\mathbf{x},y)$ for notational simplicity. In order to estimate the posterior mean for $\mathbf{g}$ given loss function values, we perturb the input using scaled version of supplied vectors $\{\mathbf{z}^{(1)}\}_{i=1}^{m}$, i.e.,  $\delta_{1}\{\mathbf{z}^{(1)}\}_{i=1}^{m}$ so as to obtain the perturbed points $\{\mathbf{x}'^{(i)}\}_{i=1}^{m}$. 
Thus, given a set of sample points ${\mathbf{x}'^{(1)},\ldots,{\mathbf{x}'}^{(m)}}$ (which here refer to different \emph{perturbations} of a single underlying point $\mathbf{x}$), corresponding loss function values $L({\mathbf{x}
'}^{(1)}),\ldots,L({\mathbf{x}'}^{(m)})$, we have the following characterization of the distribution
\begin{align*}
\mathbf{L}|\mathbf{g} \sim \mathcal{N}(\mathbf{Xg}, \sigma^2\mathbf{I}),
\end{align*}
where 
\begin{equation}
\mathbf{L} = \left [L({\mathbf{x}'}^{(1)}) - L(\mathbf{x}),\cdots, L({\mathbf{x}'}^{(m)}) - L({\mathbf{x}}) \right ], \mathbf{X} = \left [ ({\mathbf{x}'}^{(1)} - {\mathbf{x}'})^{\top}, \cdots, ({\mathbf{x}'}^{(m)} - {\mathbf{x}'})^{\top}\right ].
\label{eq:LX}
\end{equation}
With the above formalism in place, using standard Gaussian manipulation rules, the posterior of interest $\mathbf{g}|\mathbf{L}$, is given by
\begin{equation}
\mathbf{g}|\mathbf{L} \sim \mathcal{N}\left (\left (\Lambda + \mathbf{X}^{\top}\mathbf{X}/\sigma^2\right)^{-1} \mathbf{X}^{\top} \mathbf{L}_{1}/\sigma^2, \left (\Lambda + \mathbf{X}^{\top}\mathbf{X}/\sigma^2\right)^{-1} \right).
\end{equation}
To compute this estimate, we need to solve a linear equation in the matrix $\Lambda + \mathbf{X}^{\top}\mathbf{X}/\sigma^2.$ This is a convolution plus a low rank matrix (the $\mathbf{X}^{\top}\mathbf{X}$ term is rank $m$, and we typically have $m \ll n$ because we have relatively few samples and a high dimensional input). We can not solve for this matrix exactly using the FFT, but we can still solve for it efficiently (requiring only an $m \times m$ inverse) using the matrix inversion lemma, specifically using Woodburry's matrix inversion lemma
\begin{equation}
\left(\Lambda + \mathbf{X}^{\top}\mathbf{X} / \sigma^2\right )^{-1} = \Lambda^{-1} - \Lambda^{-1} \mathbf{X}^{\top} \left ( \sigma^2 \mathbf{I} + \mathbf{X}\Lambda^{-1}\mathbf{X}^{\top} \right )^{-1}\mathbf{X} \Lambda^{-1}.
\end{equation}
Since the term needs to be computed explicitly for at least the inner inverse, we explicitly maintain the term $\mathbf{U} = \Lambda^{-1}\mathbf{X}^{\top}$.  Note that in the sequential sampling setting (where we sequentially sample $\mathbf{x}^{(i)}$ points one at a time), this matrix can be maintained over all samples, and the factorization updated sequentially.
\begin{algorithm}[t]
\caption{GMRF based black-box FGSM}
\label{alg:bbfgsm}
\begin{algorithmic}[1]
\Procedure{\textsc{BB-FGSM}}{$\sigma,\delta, \epsilon, \theta$}
  \State Construct perturbed points: ${\mathbf{x}'}^{(1:m)} := \mathbf{x} + \delta \cdot \textrm{Top FFT basis vectors}$
  \State Form $\mathbf{L}$ and $\mathbf{X}$ using \eqref{eq:LX}
  \State Using FFT, compute: $\mathbf{\widehat{g}} := \left (\Lambda(\theta) + \mathbf{X}^{\top}\mathbf{X}/\sigma^2\right)^{-1} \mathbf{X}^{\top} \mathbf{L}/\sigma^2$
  \State \Return $\mathbf{x} + \epsilon \cdot \mathrm{sign}(\hat{\mathbf{g}})$
\EndProcedure
\end{algorithmic}
\end{algorithm}

Finally, we use this posterior inference within a simple FGSM-like attack method for $\ell_\infty$ attacks, which first computes the mean estimate of the gradient $\hat{\mathbf{g}}$, then constructs an adversarial example in the direction of its sign.  One free parameter of the method is how we choose the perturbations to create ${\mathbf{x}'}^{(1:m)}$.  While we could use simple Gaussian perturbations, our covariance structure strongly motivates an alternative choice: using the top FFT basis vectors, which correspond to the dominant eigenvalue of the Covariance, and thus given the directions of highest expected variance.  While FFT vectors have been used previously in adversarial attacks \citep{yin2019fourier}, our formulation thus further provides a motivation for their adoption.  The full method is shown in Algorithm \ref{alg:bbfgsm}.

\paragraph{A note on the threat model.} One important point to note is that, unlike typical black box models, the \textsc{BB-FGSM} algorithm implicitly uses some information from \emph{multiple} images to form it's attack (because we estimate the GMRF parameters $\theta$ on separate images, using additional queries to the classifier).  However, this is a relatively minor difference in practice: for a given classifier we estimate $\theta$ \emph{once} offline, and because $\theta$ has so few parameters (typically just 3--4), we need a very small number of total queries to the classifier to estimate it (e.g., we use 10 additional images with 10 queries each to estimate the parameters for ImageNet classifiers).  After learning these parameters offline, they can be used indefinitely to run attacks against the classifier, essentially amortizing away any computational cost of estimating the parameters.

In order to achieve query efficiency our framework incorporates gradient correlation into gradient estimation through the GMRF framework which we then further exploit by the procedure described above.
In order to estimate the adversarial direction efficiently in Algorithm \ref{alg:gradest}, a key role is played by the vectors $\{\mathbf{z}^{(1)}\}_{i=1}^{m}$ which perturb the input. One particular choice being sampling the directions from a normal distribution. 
However, it is worth noting that the inverse covariance distribution of the gradients by construction is a convolution operator and hence is diagonalized by the FFT basis. 
Low frequency components of the FFT basis vectors enhanced the attack accuracy significantly, an example of which is depicted in Figure \ref{fig:0} for black-box attacks on VGG16-bn classifier for ImageNet with the $\ell_\infty$ perturbation constrained to $\epsilon=0.05$.
The adversarial input for $\mathbf{x}$ is generated using FGSM as follows:
 \begin{align}
     \label{eq:bb-fgsm}
     \mathbf{x}_{adv}=\mathbf{x}+\epsilon\textrm{sign}\left(\mathbf{\widehat{g}}\right).
\end{align}
The entire procedure consisting of GMRF inference and gradient inference is presented in Algorithm \ref{alg:bbfgsm}. We defer the discussion of computing FFT basis vectors to the appendix.

\section{Experiments}
In this section, we evaluate our proposed gradient estimation and resulting black box attack on the MNIST \citep{mnist} and ImageNet \citep{imagenet} data sets, considering untargeted $\ell_\infty$ attacks in particular.  We evaluate the performance of our approach against several alternative gradient-based black-box attacks, in terms of attack success rate\footnote{The attack success rate is defined as the ratio of the number of images successfully misclassified to the total number of input images. Among the set of input images, we discard images which were misclassified by the classifier. The average query count is computed on the queries made for successful attacks.} with a given query budget and average query count (we don't compare to non-gradient-based methods, as the overall evaluation here is to emphasize the importance of gradient estimation in the low-query setting, not necessarily to demonstrate the absolute most success black-box attacks).  More complete experimental details are presented in the appendix. The code is available at \url{https://github.com/anitksahu/GMRF}.
\begin{figure*}[t]
\label{fig:1}
    \centering
     \includegraphics[scale=0.16]{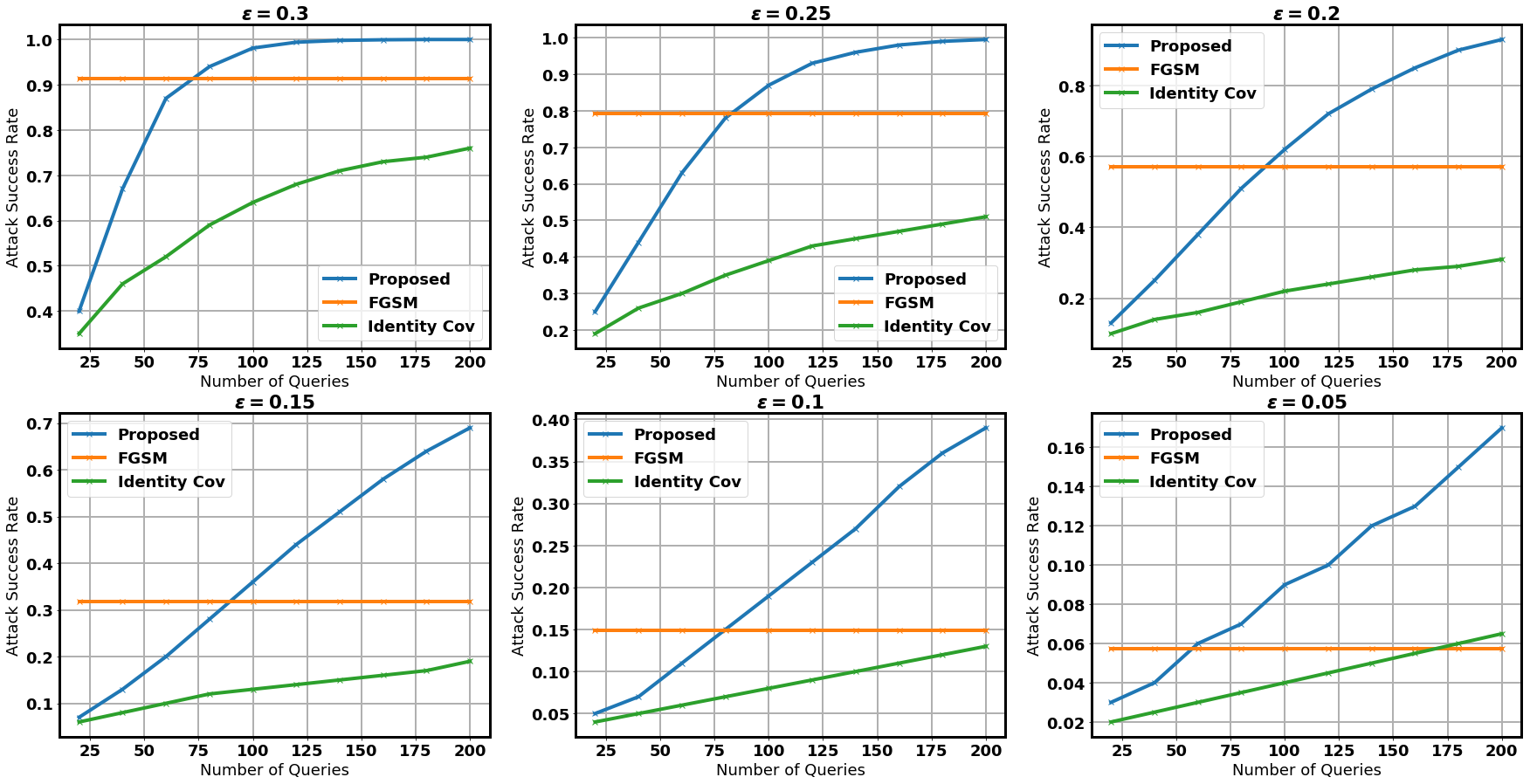}
     \caption{Performance comparison of the proposed attack, the proposed attack with identity gradient covariance and white-box FGSM in terms of attack accuracy with query budgets between 20 and 200 for MNIST on LeNet.}
 \end{figure*}

\begin{figure}
\centering
\begin{subfigure}[b]{0.8\textwidth}
   \includegraphics[width=\linewidth]{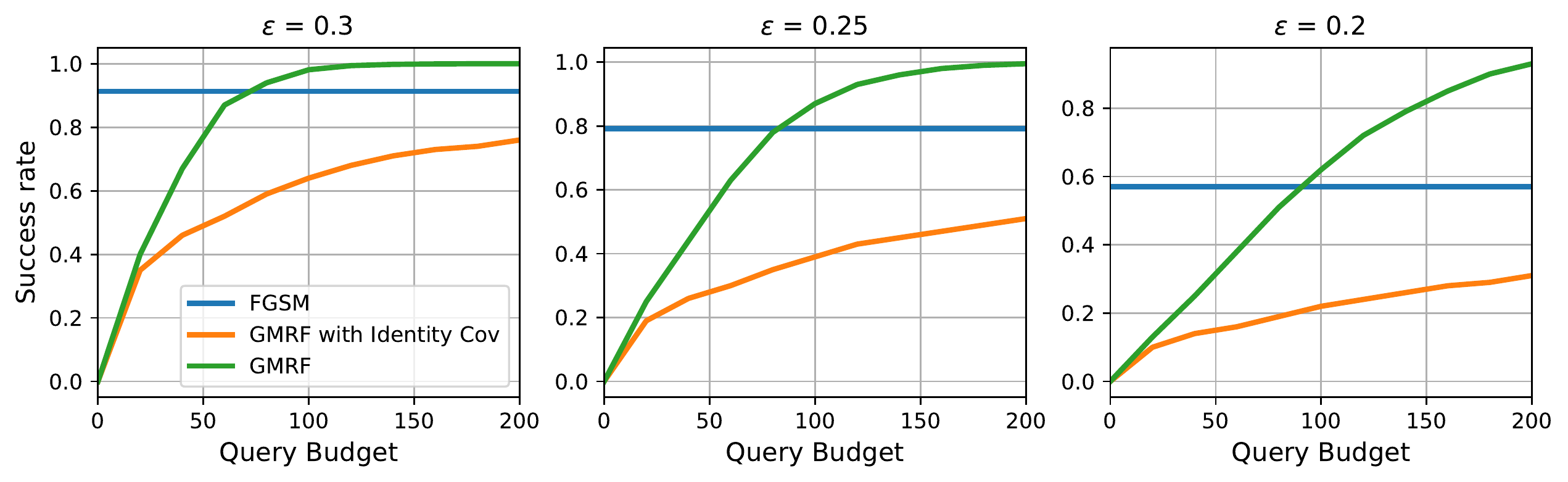}
\end{subfigure}
\begin{subfigure}[b]{0.8\textwidth}
   \includegraphics[width=1\linewidth]{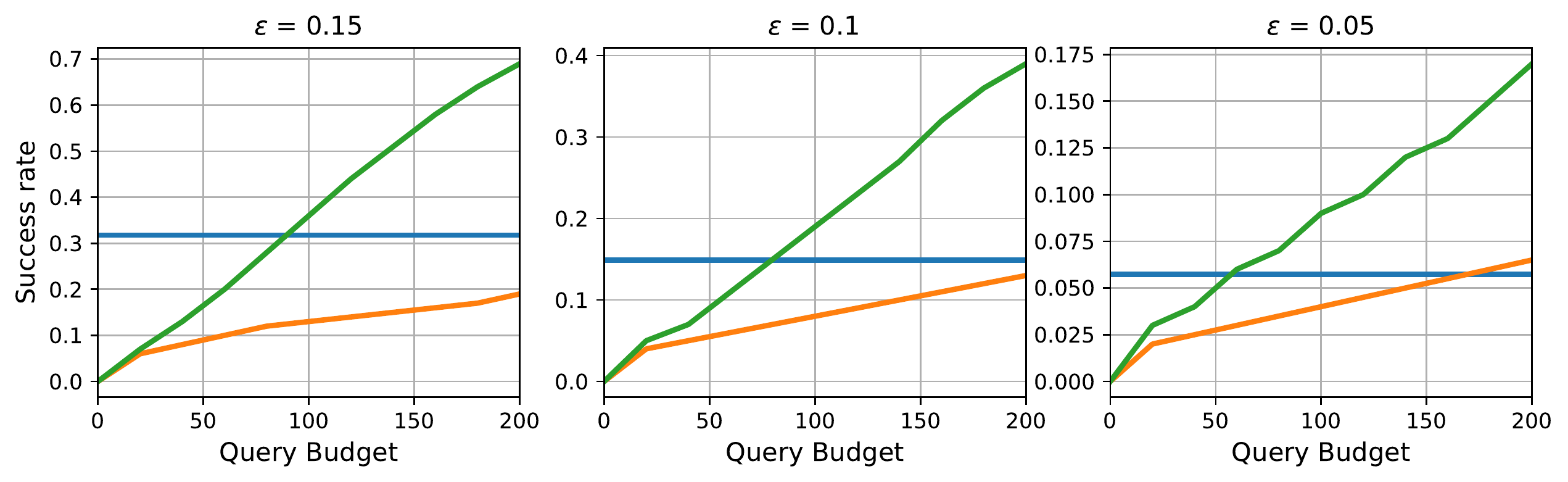}
\end{subfigure}
\caption{Performance comparison of the proposed attack (GMRF), the proposed attack with identity gradient covariance and white-box FGSM in terms of attack accuracy with query budgets between 20 and 200 for MNIST on LeNet.}
\label{fig:1}
\end{figure}

\subsection{Experiments on MNIST}
We first compare the performance of the proposed method with that of white-box FGSM \citep{goodfellow2014explaining} across different values of $\ell_\infty$ bounds ranging from $0.05$ to $0.3$ in increments of $0.05$ over query budgets from $20$ to $200$. To further illustrate the importance of incorporating a non-identity gradient covariance, we also provide experimental results for a version of our proposed algorithm which takes the gradient covariance to be an identity matrix.  

As shown in Figure 4, our proposed attack method exhibits better attack accuracy than white-box FGSM in around $75$ queries and the gap in the performance is further magnified with increasing number of queries. On the other hand, the version of our algorithm with identity gradient covariance consistently under performs with respect to the white-box FGSM attack. This further illustrates the effectiveness of our proposed algorithm and the importance of modelling the gradient covariance.  We defer additional discussions concerning cosine similarities of the adversarial directions of the proposed GMRF based attack framework and white-box FGSM to the appendix.

The superior performance of our proposed framework as compared to white-box FGSM as demonstrated in Figure 4 can be attributed to the following reason.  First, incorporating the gradient non-identity covariance structure into the gradient estimation scheme, allows our perturbation to be able to use structural gradient information from other images too. On the other hand, white box FGSM treats gradient of every image to be independent of gradients of other images.  
This is further illustrated by our experimental findings based on the version of our algorithm considering the gradient covariance to be an identity matrix.

\subsection{Experiments on ImageNet}
\begin{figure*}[t]
	\centering
	\minipage{0.32\textwidth}
	\includegraphics[scale=0.4]{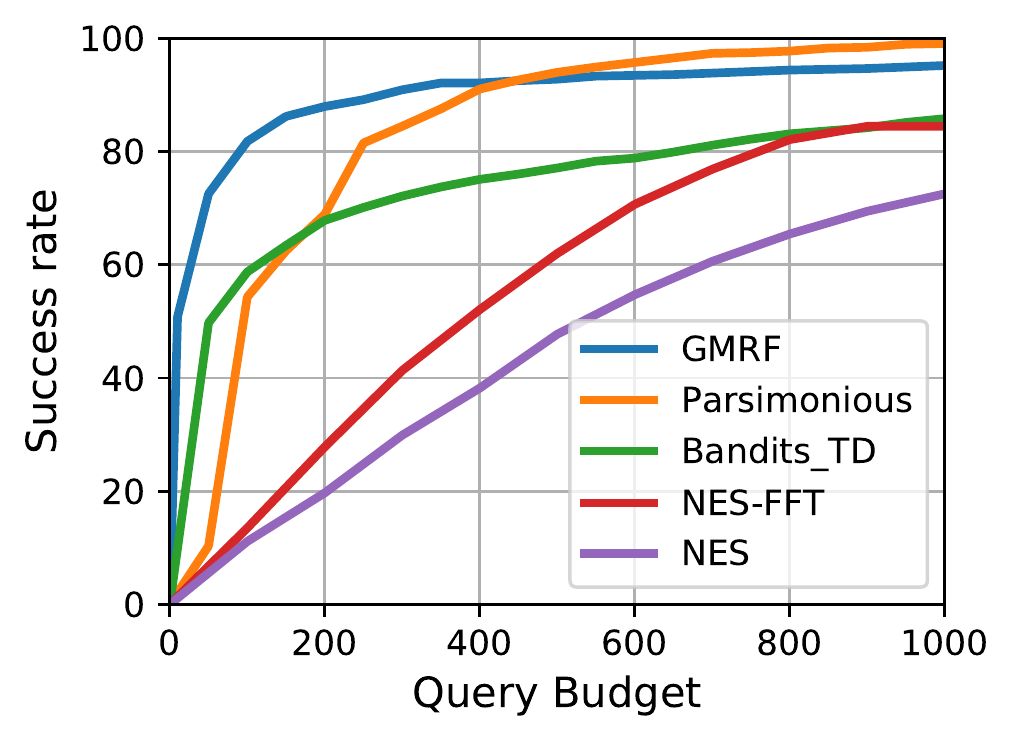}
	\caption{VGG16-bn: Attack success rate as a function of query budget}
	\centering
	\label{fig:vgg}
	\endminipage\hfill
	\minipage{0.32\textwidth}
	\includegraphics[scale=0.4]{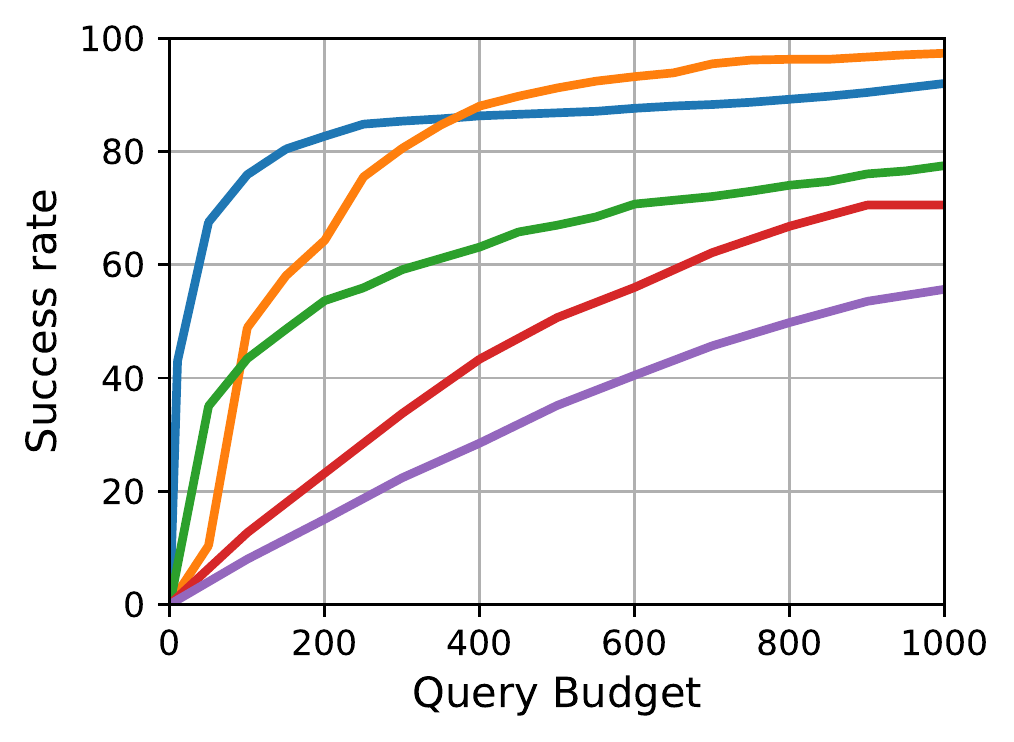}
	\caption{ResNet50: Attack success rate as a function of query budget}
	\centering
	\label{fig:resnet}
	\endminipage\hfill
	\minipage{0.32\textwidth}
	\includegraphics[scale=0.4]{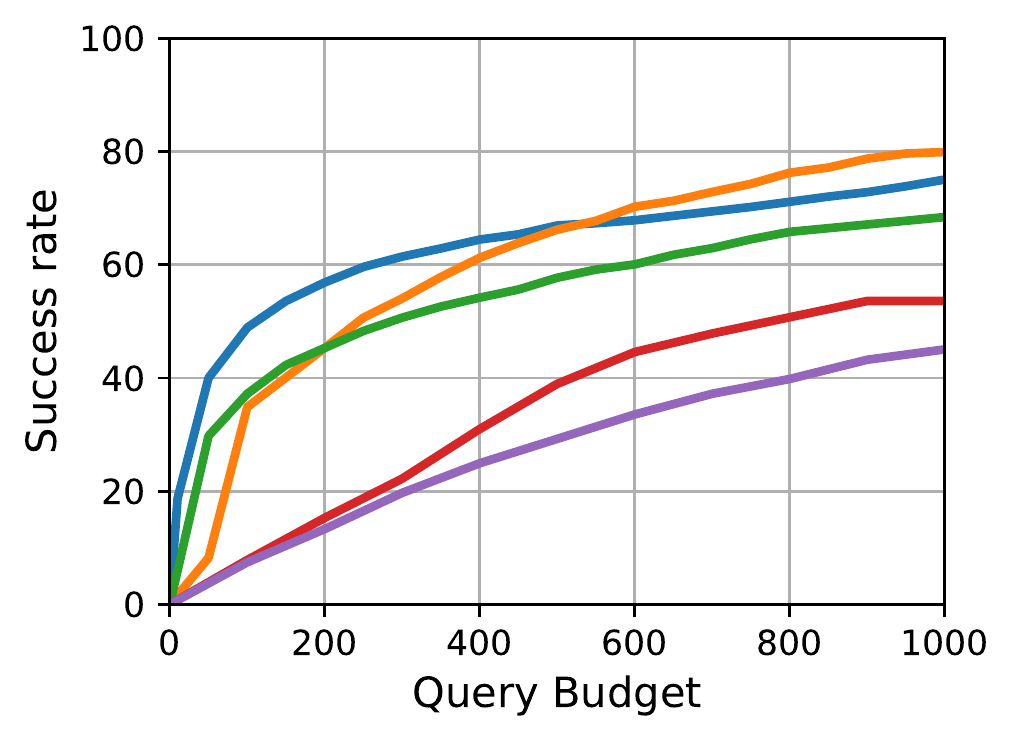}
	\caption{Inception-v3: Attack success rate as a function of query budget}
	\centering
	\label{fig:inception}
	\endminipage
	%\minipage{0.32\textwidth}
	%\includegraphics*[viewport=0 0 500 430, width = 2.05in, height = 2.05in]{example-image-a}
	%\centering
	%\endminipage
	%\caption{Attack success rate as a function of query budget}
	%	\vspace*{-10pt}
\end{figure*}
\begin{table*}[t]
\centering
\caption{Summary of $\ell_\infty$ attacks with $\epsilon=0.05$ ImageNet
attacks using \textsc{NES}, \textsc{Bandits-td}, \textsc{Parsimonious} and the proposed method with a query budget of 1000 per image
}

\label{tab:multistep}
  \begin{tabular}{@{}ccccccccc@{}}
    \toprule
    \multirow{2}{*}{\textbf{Attack}} & 
    \multicolumn{2}{c}{\textbf{VGG16-bn}} & \phantom{x} &
    \multicolumn{2}{c}{\textbf{Resnet50}} & \phantom{x} &
    \multicolumn{2}{c}{\textbf{Inception-v3}} \\ 
    \cmidrule{2-3} \cmidrule{5-6} \cmidrule{8-9} & Success &
    Avg. Queries && Success &
    Avg. Queries && Success &
    Avg. Queries \\
    \midrule
    NES & 72.52\% & 452.86 && 55.65\% & 463.63 && 45.03\% & 448.40\\
    NES-FFT & 84.43\% & 395.10 && 70.53\% & 405.86 && 53.59\% & 417.80\\
    Parsimonious & \bf{99.06}\% & 165.47  && \bf{97.33}\% & 178.61 && \bf{79.89}\% & 253.95 \\
    Bandits$_{TD}$ & 85.79\% & 141.16 &&
    77.49\% &  190.47 && 68.40\% & 204.06 \\
    Proposed & 95.17\% & \bf{56.79} && 92.01\% & \bf{84.06} && 75.03\% & \bf{170.09} \\
    \bottomrule
  \end{tabular}
\end{table*}
We next compare the performance of the proposed method on ImageNet with that of \textsc{NES} \citep{nes}, \textsc{Bandits-td} \citep{bandits}, a version of \textsc{NES} using FFT basis vectors and \textsc{Parsimonious} \citep{parsimonious}, which are the current state of the art in $\ell_\infty$ gradient-based black-box attacks. We consider three pre-trained classifiers namely, ResNet50 \citep{resnet}, Inception-v3 \citep{inception} and VGG16-bn \citep{vgg}. We use the pre-trained models provided by PyTorch for attacking these classifiers. We evaluation on a randomly sampled set of $1000$ images from ImageNet validation set. 

The $\ell_\infty$ perturbation bound is set to $\epsilon=0.05$ (with images noramlized to [0,1]. We use the implementation and hyperparameters provided by \citet{bandits} for \textsc{NES} and \textsc{Bandits-td}. Similarly for \textsc{Parsimonious}, we use the implementation and hyperparameters given by \citep{parsimonious}.The specifics of the GMRF model, the values of the parameters and the associated hyperparameters (which were obtained by grid search) for the proposed algorithm for the three classifiers are relegated to the Appendix. 

Figures \ref{fig:vgg} - \ref{fig:inception} show the evolution of attack accuracy with different querying budgets (the amortized number of queries needed to estimate GMRF parameters are also added to the totals for the GMRF method, but this adds an average of only $0.2$ queries per example). In the low query budget setting, with less than $450$ query budget, our algorithm outperforms \textsc{parsimonious}, though it exhibits slightly inferior performance with higher queries. 
As shown in Table \ref{tab:multistep}, our proposed algorithm, in spite of being a single-step attack, outperforms \textsc{Bandits-td}, \textsc{NES-FFT} and \textsc{NES} by achieving higher attack success rate. Despite the higher success rate, the proposed method uses much fewer queries on an average as compared to \textsc{Bandits-td}, \textsc{NES-FFT} and \textsc{NES}. Thus, the proposed method strictly dominates \textsc{Bandits-td}, \textsc{NES-FFT} and \textsc{NES} on every metric. 
 
The proficiency of our proposed scheme in the low query budget regime can be attributed to the utilization of the GMRF framework and the usage of FFT basis vectors. In particular, the FFT basis vectors being the eigen vectors of the covariance matrix provide for systematic dimensionality reduction.

\section{Conclusion}
In this paper, we proposed an efficient method for modeling the covariance of gradient estimated in deep networks using Gaussian Markov random fields.  Although the method naively would be  computationally impractical, we show that by employing an FFT approach, we can efficiently learn and perform inference over the Gaussian covariance structure.  We then use this approach to highlight an efficient gradient-based black-box adversarial attack, which, owing to capturing this correlational structure, typically requires fewer samples than existing gradient-based techniques.

Although the highlight of this work involved using our gradient estimation procedure within a black box adversarial attack, the proposed methods also have applicability in any setting where one wishes to obtain an estimate of neural network input gradients, using limited queries to the model.  We hope to explore further applications and algorithmic advances using these techniques is future work.

\section{Broader Impacts}
The sensitivity of deep learning classifiers to adversarial attacks raises serious concerns about their practical application in many real-world, safety-critical domains.  Black box attacks, in particular, highlight the potential of these attacks to apply to practically deployed systems, where we do not have access to the full classifier definition.  At the same time, the methods proposed here are not easily applied immediately to many deployed classifiers in the real world, as they require multiple classifications of the same input with different perturbations, and also require access to the loss.  Thus, while this is a step toward demonstrating the feasibility of real-world attacks, there are still obstacles to overcome if adversaries wished to deploy them in real-world settings.

Nonetheless, the shift towards more black-box versions of adversarial attacks continues to pose a major challenge for deployed systems, which should be constantly evaluated.  Our hope is that the presence of effective attacks makes it clear the limitations of such systems, and ultimately encourages the development of more robust models to deploy in the real world.

% In this paper, we have developed a GMRF based covariance modeling technique so as to streamline the gradient estimation scheme catered towards black-box adversarial attacks. In particular, due to the streamlined gradient estimation scheme incorporating correlations, we could alleviate the issue of random direction based searches, which plagues every zeroth order optimization scheme due to biased gradient estimates. The gradient estimation scheme can be used in any gradient based black-box adversarial attack method to attain higher attack accuracies with lower query counts. Our method facilitates single iteration based query efficient black-box attacks which we demonstrated to be as potent as multi-step attacks on multiple architectures and datasets in terms of attack success rate. We also employed techniques from matrix analysis and FFT to make our attack computationally efficient. Our results open avenues for more effective covariance modeling techniques so as to further streamline gradient estimation schemes so as to facilitate more query efficient black-box adversarial attacks.

\bibliographystyle{abbrvnat}
\bibliography{iclr2020_conference}

\newpage
\appendix
\section{Appendix}

\subsection{Computation of FFT basis}
We use the fact that the covariance and the inverse covariance matrix because of being convolutional operators are diagonalized by the FFT basis. Let us assume the image is of size $c\times h \times w$, where $c$,$h$ and $w$ denote the number of channels, height and width of the image. We define a tensor $S$ of zeros of size $c\times h \times w \times 2$, where the last dimension is to account for both the real and complex components. In order to generate the lowest frequency basis vector, we set the the first element of the tensor of the first channel, i.e, $S_{0,0,0,0}=1$ and take the inverse FFT. This gives us the lowest cosine basis vector. We do the same for the other channels, by just setting the corresponding component to $1$ and taking the inverse FFT. Setting, $S_{0,0,0,1}=1$ and then taking the inverse FFT yields the lowest frequent sine component. In order to generate the low frequency components, we start from the beginning of a row and proceed along diagonally by incrementing the row and column index by one. At each entry of the tensor, we repeat it for every channel once at a time.

\subsection{Comparison of effectiveness of directions}
\begin{figure}
    \centering
   \includegraphics[scale=0.2]{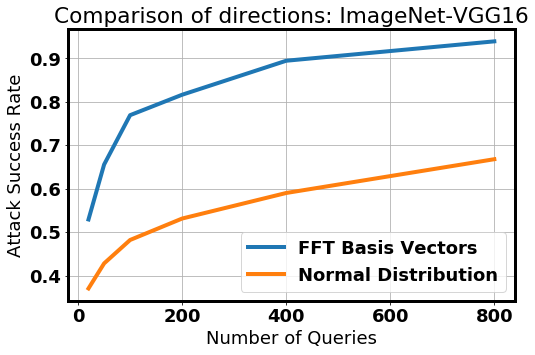}
     \caption{FFT basis vectors vs normal distribution for VGG16-bn.}
    \label{fig:0}
\end{figure}
In order to estimate the adversarial direction efficiently in Algorithm \ref{alg:gradest}, a key role is played by the vectors $\{\mathbf{z}^{(1)}\}_{i=1}^{m}$ which perturb the input. One particular choice being sampling the directions from a normal distribution. 
However, it is worth noting that the inverse covariance distribution of the gradients by construction is a convolution operator and hence is diagonalized by the FFT basis. 
Low frequency components of the FFT basis vectors enhanced the attack accuracy significantly, an example of which is depicted in Figure \ref{fig:0} for black-box attacks on VGG16-bn classifier for ImageNet with the $\ell_\infty$ perturbation constrained to $\epsilon=0.05$.

\subsection{MNIST Experiments}
The GMRF model used for MNIST is given by $\Lambda_{i,i}=\alpha, \Lambda_{i,i+1}=\Lambda_{i+1,i}=\Lambda_{i,i-1}=\Lambda_{i-1,i}=\beta, \Lambda_{i+1,i+1} = \Lambda_{i-1,i-1}= \Lambda_{i-1,i+1}= \Lambda_{i+1,i-1} = \gamma$, where $\Lambda_{i,j}$ denotes the $(i,j)$-th element of $\Lambda$. For estimating the GMRF parameters, we use the last $5$ images of the MNIST test set and perturb each of them with $5$ vectors drawn from a normal distribution. For the attack, we use low frequency basis vectors of the FFT basis. The following table gives the values of the different hyperparameters used in the attack.
\begin{table}[ht]
\centering
\caption{MNIST Experiment Settings}
\begin{tabular}{|c|c|}
\toprule
    $\delta$ & $0.1$ \\
    \midrule
    $\alpha$ & $21094408$\\
    \midrule
    $\beta$  & $-5116365$\\
    \midrule
    $\gamma$  & $284558.1562$\\
    \midrule
    $\sigma$ & $10^{-3}$\\
    \midrule
    $\delta_1$ & $0.15$\\
\bottomrule
\end{tabular}
\end{table}
Except for the GMRF parameters, all the other parameters were determined using grid search.
\subsection{VGG16 Experiments}
The GMRF model used for VGG16 for Imagenet is given by $\Lambda_{0,i,i}=\alpha, \Lambda_{0,i,i+1}=\Lambda_{0,i+1,i}=\Lambda_{0,i,i-1}=\Lambda_{0,i-1,i}=\beta$, $\Lambda_{0,i+1,i+1} = \Lambda_{0,i-1,i+1} = \Lambda_{0,i-1,i-1} = \Lambda_{0,i+1,i-1} = \kappa$, $\Lambda_{1,i,i}=\Lambda_{-1,i,i}=\gamma$, where in $\Lambda_{k,i,j}$, $k$ denotes the channel. We also tried out GMRF models of lower and higher degree of association and we selected the one performing the best. For estimating the GMRF parameters, we use the last $10$ images of the ImageNet validation set and perturb each of them with $10$ vectors drawn from a normal distribution. For the attack, we use low frequency cosine basis vectors of the FFT basis. The following table gives the values of the different hyperparameters used in the attack.
\begin{table}[ht]
\centering
\caption{ImageNet VGG-16 Experiment Settings}
\begin{tabular}{|c|c|}
\toprule
    $\delta$ & $1.0$ \\
    \midrule
    $\alpha$ & $633.44$\\
    \midrule
    $\beta$  & $-24.05$\\
    \midrule
    $\gamma$  & $-232.04$\\
    \midrule
    $\kappa$  & $-2.00$\\
    \midrule
    $\sigma$ & $1.0$\\
    \midrule
    $\delta_1$ & $0.04$\\
\bottomrule
\end{tabular}
\end{table}
Except for the GMRF parameters, all the other parameters were determined using grid search.
\subsection{ResNet50 Experiments}
The GMRF model used for VGG16 for Imagenet is given by $\Lambda_{0,i,i}=\alpha, \Lambda_{0,i,i+1}=\Lambda_{0,i+1,i}=\Lambda_{0,i,i-1}=\Lambda_{0,i-1,i}=\beta$, $\Lambda_{0,i+1,i+1} = \Lambda_{0,i-1,i+1} = \Lambda_{0,i-1,i-1} = \Lambda_{0,i+1,i-1} = \kappa$, $\Lambda_{0,i,i+2} = \Lambda_{0,i,i-2} = \Lambda_{0,i-2,i} = \Lambda_{0,i+2,i} = \Lambda_{0,i+1,i+2}=\Lambda_{0,i-1,i+2}=\Lambda_{0,i+2,i+1}=\Lambda_{0,i+2,i-1}=\Lambda_{0,i-1,i-2}=\Lambda_{0,i+1,i-2}=\Lambda_{0,i-2,i-1}=\Lambda_{0,i-2,i+1}=\nu$, $\Lambda_{1,i,i}=\Lambda_{-1,i,i}=\gamma$, where in $\Lambda_{k,i,j}$, $k$ denotes the channel. We also tried out GMRF models of lower and higher degree of association and we selected the one performing the best. For estimating the GMRF parameters, we use the last $10$ images of the ImageNet validation set and perturb each of them with $10$ vectors drawn from a normal distribution. For the attack, we use low frequency cosine basis vectors of the FFT basis. The following table gives the values of the different hyperparameters used in the attack.
\begin{table}[ht]
\centering
\caption{ImageNet ResNet50 Experiment Settings}
\begin{tabular}{|c|c|}
\toprule
    $\delta$ & $1.0$ \\
    \midrule
    $\alpha$ & $2631.93$\\
    \midrule
    $\beta$  & $-263.33$\\
    \midrule
    $\gamma$  & $-837.16$\\
    \midrule
    $\kappa$  & $6.78$\\
    \midrule
    $\nu$  & $28.09$\\
    \midrule
    $\sigma$ & $0.5$\\
    \midrule
    $\delta_1$ & $0.0375$\\
\bottomrule
\end{tabular}
\end{table}
Except for the GMRF parameters, all the other parameters were determined using grid search.
\subsection{Inception v3 Experiments}
The GMRF model used for VGG16 for Imagenet is given by $\Lambda_{0,i,i}=\alpha, \Lambda_{0,i,i+1}=\Lambda_{0,i+1,i}=\Lambda_{0,i,i-1}=\Lambda_{0,i-1,i}=\beta$, $\Lambda_{0,i+1,i+1} = \Lambda_{0,i-1,i+1} = \Lambda_{0,i-1,i-1} = \Lambda_{0,i+1,i-1} = \kappa$, $\Lambda_{0,i,i+2} = \Lambda_{0,i,i-2} = \Lambda_{0,i-2,i} = \Lambda_{0,i+2,i} = \Lambda_{0,i+1,i+2}=\Lambda_{0,i-1,i+2}=\Lambda_{0,i+2,i+1}=\Lambda_{0,i+2,i-1}=\Lambda_{0,i-1,i-2}=\Lambda_{0,i+1,i-2}=\Lambda_{0,i-2,i-1}=\Lambda_{0,i-2,i+1}=\nu$, $\Lambda_{1,i,i}=\Lambda_{-1,i,i}=\gamma$, where in $\Lambda_{k,i,j}$, $k$ denotes the channel. We also tried out GMRF models of lower and higher degree of association and we selected the one performing the best. For estimating the GMRF parameters, we use the last $10$ images of the ImageNet validation set and perturb each of them with $10$ vectors drawn from a normal distribution. For the attack, we use low frequency cosine basis vectors of the FFT basis. The following table gives the values of the different hyperparameters used in the attack.
\begin{table}[ht]
\centering
\caption{ImageNet Inception v3 Experiment Settings}
\begin{tabular}{|c|c|}
\toprule
    $\delta$ & $1.0$ \\
    \midrule
    $\alpha$ & $8964.89$\\
    \midrule
    $\beta$  & $-2960.87$\\
    \midrule
    $\gamma$  & $-841.13$\\
    \midrule
    $\kappa$  & $1155.66$\\
    \midrule
    $\nu$  & $286.03$\\
    \midrule
    $\sigma$ & $0.1$\\
    \midrule
    $\delta_1$ & $0.045$\\
\bottomrule
\end{tabular}
\end{table}
Except for the GMRF parameters, all the other parameters were determined using grid search.

\subsection{Gradient Estimation Performance}
\begin{figure}[t]
\centering
\minipage{0.45\textwidth}
  \includegraphics[scale=0.45]{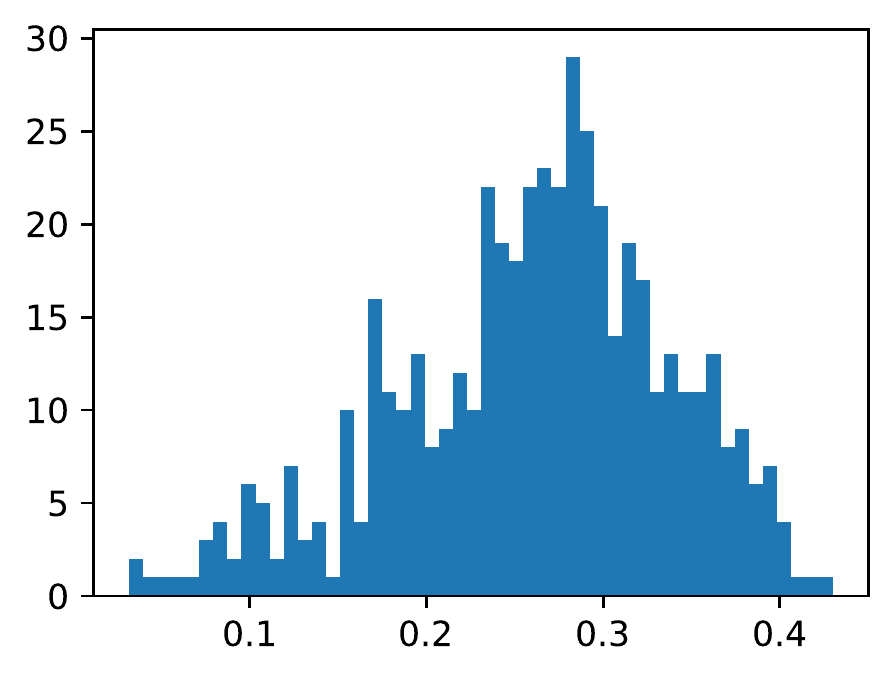}
  \caption{Cosine Similarity MNIST: 40 queries}
  \label{fig:sim_mnist_40_f}
	\endminipage\hfill
\minipage{0.45\textwidth}
   \includegraphics[scale=0.45]{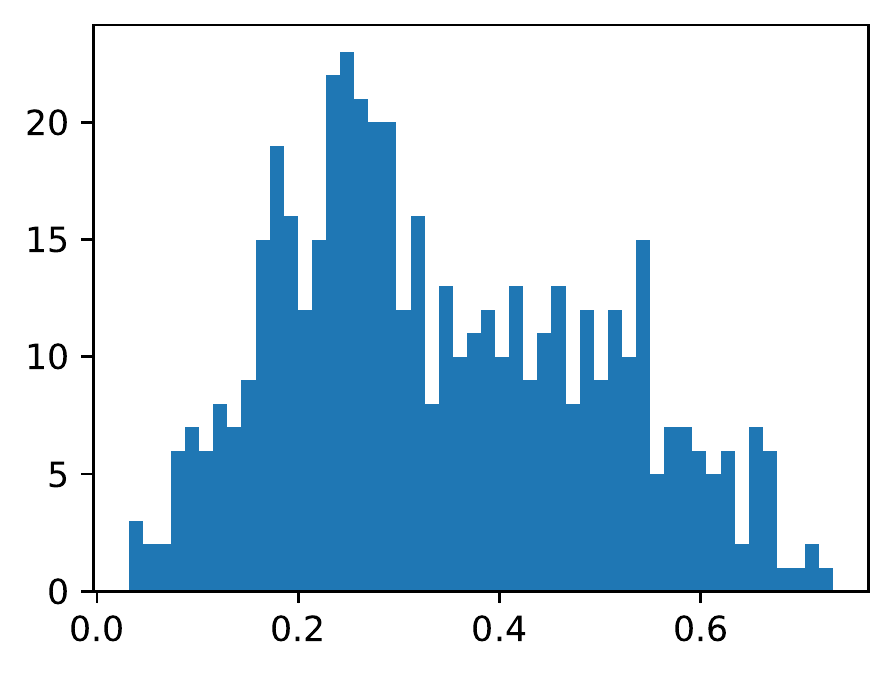}
   \caption{Cosine Similarity MNIST: 200 queries}
   \label{fig:sim_mnist_all_f}
\endminipage
\end{figure}

We explore the cosine similarities of the adversarial directions of the proposed GMRF based attack framework and white-box FGSM. For the $40$ query budget setting, the cosine similarity is centered around $0.3$, while for the $200$ query budget setting, the cosine similarity is centered around $0.2$ as depicted in figures \ref{fig:sim_mnist_40_f} and \ref{fig:sim_mnist_all_f} respectively. Had our gradient estimates completely aligned with that of the true gradient as in white-box FGSM our performance would have been upper bounded by the performance of white-box FGSM. In essence, our scheme finds directions for adversarial perturbations, which in itself does not maximize the loss but is able to find a direction which leads to misclassification of the examples and outperform white-box FGSM. 
We further illustrate the performance of our gradient estimation scheme through experiments on the ImageNet dataset using the VGG-16bn architecture. We use two metrics namely, mean squared error of the normalized estimated gradient with respect to the normalized true gradient and the  cosine similarity between the estimated gradient and the true gradient. In order to perform our analysis, we use $500$ data samples from the test set to estimate the gradient using the GMRF framework. First, we estimate the GMRF parameters as previously described in Algorithm 1 and then perform the MAP estimation for the gradient. 

The gradient estimation performance for ImageNet using VGG16-bn is depicted in Figure 8. In order to perform our analysis, we sample $500$ data samples from the ImageNet validation set to estimate the gradient using the GMRF framework. First, we estimate the GMRF parameters as previously described in Algorithm 1 and then perform the MAP estimation for the gradient. We specifically consider the query budget to be $200$. Out of the $400$ correctly classified images, white-box FGSM and our proposed algorithm attain attack accuracies of $0.9268$ and $0.8794$ respectively. While, the gradient estimation performance in terms of MSE is impressive, the cosine similarity shows that the estimated gradient does not quite coincide directionally with the true gradient. The difference in the directions explains the inferior performance of our proposed scheme in this regime. It is worth noting that the dimension of the input data for VGG16-bn is $150528$. From classical results in zeroth-order optimization it is well known that in a $d$-dimensional space, $O(d)$ queries are required to obtain a nearly bias-free gradient estimate. Our framework uses only $200$ queries to estimate the gradient which resides in a $150528$ dimensional space. In spite of the possible erroneous directional characteristic of our estimated gradient, it still manages to achieve a $0.87$ success rate and outperforms \textsc{Bandits-td} and \textsc{Parsimonious} in the $200$ query budget regime.

\begin{figure}[!htb]
\centering
\begin{subfigure}[b]{0.34\textwidth}
   \includegraphics[width=1\linewidth]{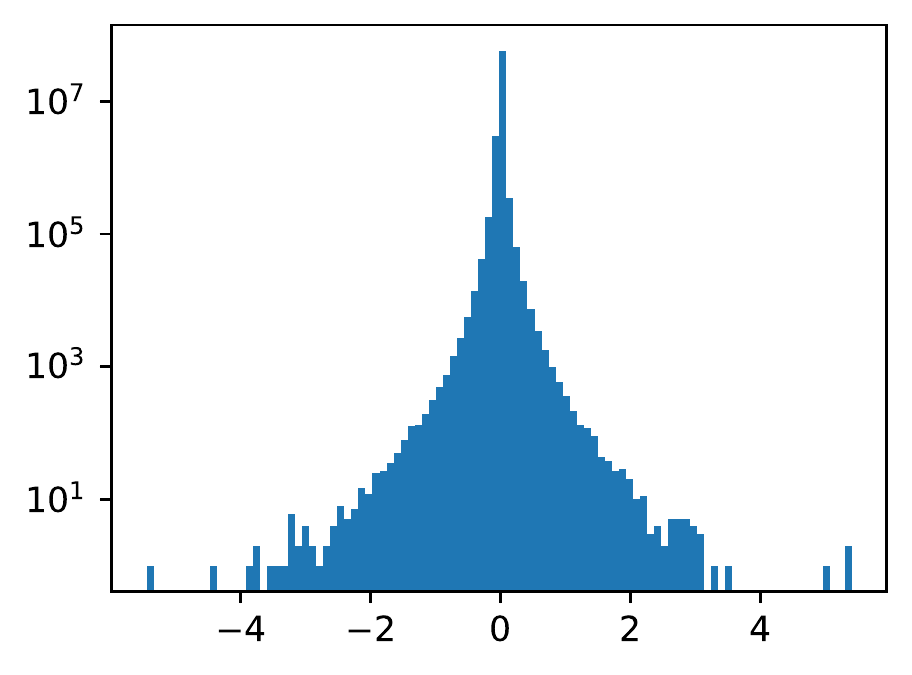}
   \caption{True Gradient Distribution}
   \label{fig:grad_dist_vgg_200}
\end{subfigure}
\begin{subfigure}[b]{0.32\textwidth}
   \includegraphics[width=1\linewidth]{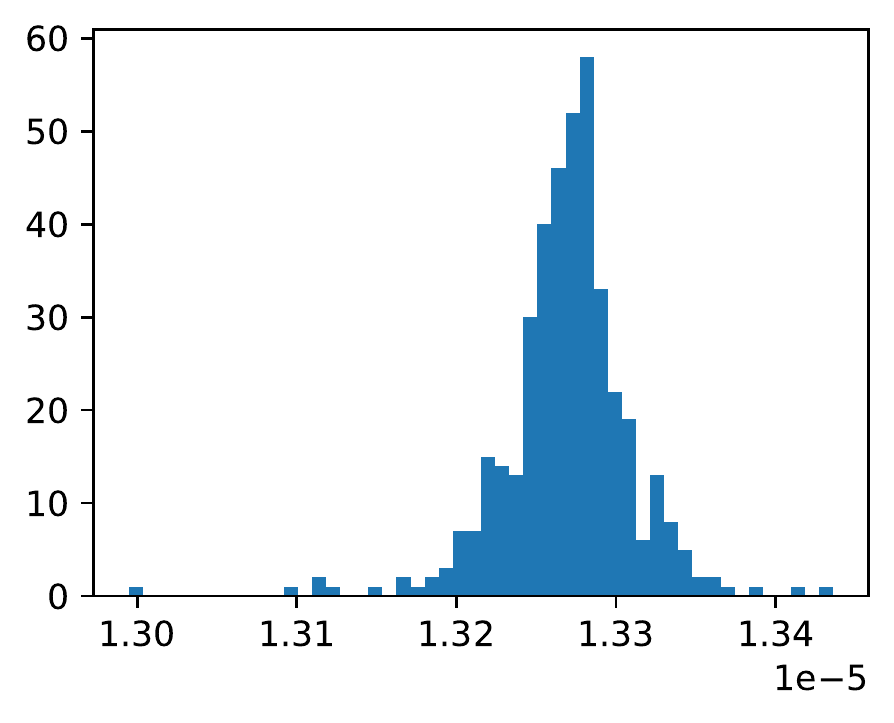}
   \caption{Mean squared error}
   \label{fig:mse_grad_vgg_200}
\end{subfigure}
\begin{subfigure}[b]{0.32\textwidth}
   \includegraphics[width=1\linewidth]{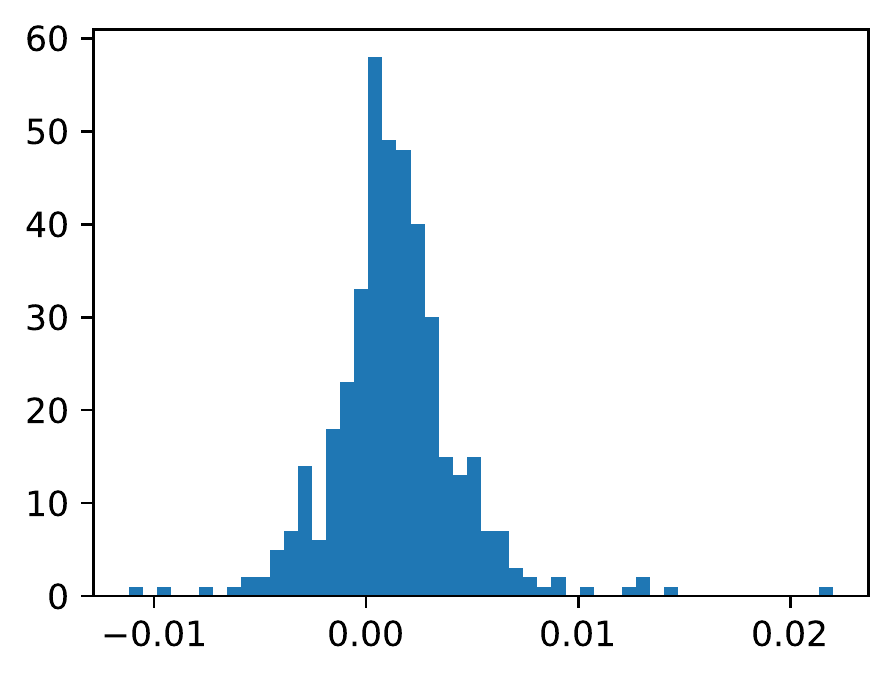}
   \caption{Cosine Similarity}
   \label{fig:sim_vgg_200}
\end{subfigure}
\caption{VGG-16bn, ImageNet, 200 query budget}
\label{fig:imagenet}
\end{figure}
\subsection{Autocorrelation}
\label{subsec:auto_corr}
\begin{figure}[h]
\centering
\includegraphics[width=0.3\linewidth]{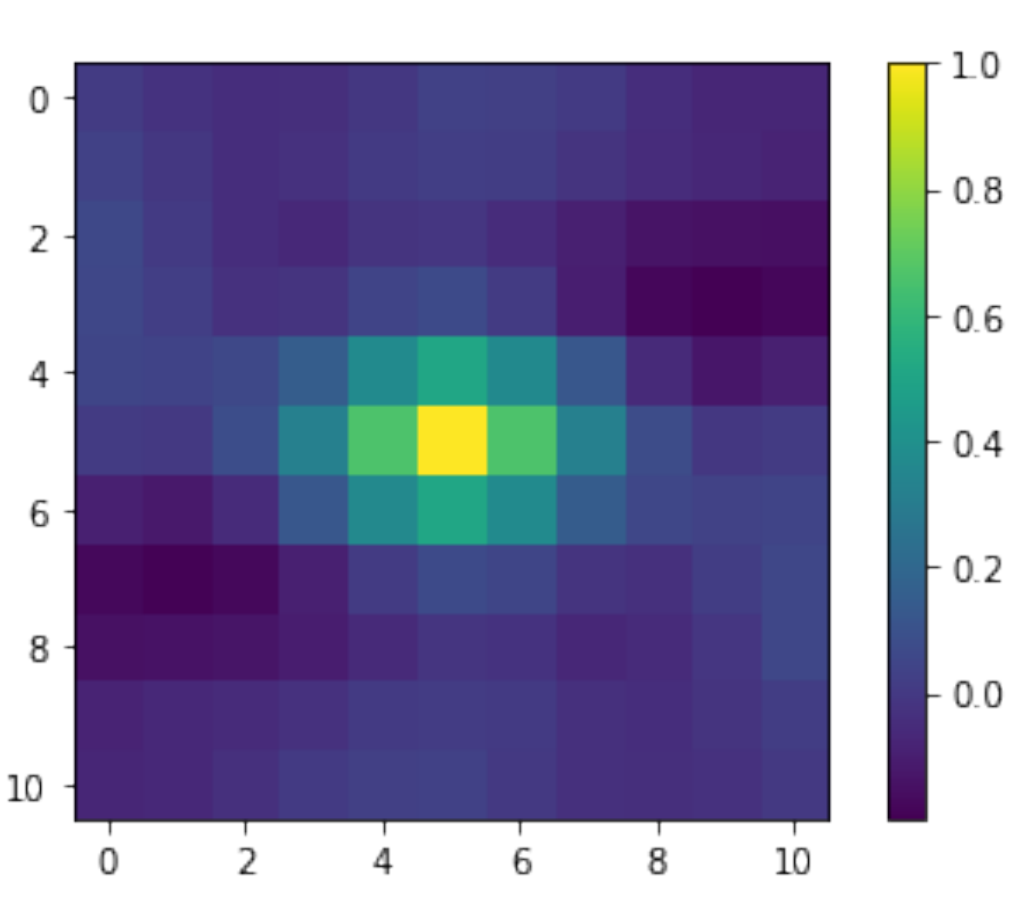}
\caption{Autocorrelation of the gradients: MNIST, LeNet, $11\times 11$ kernel}
\label{fig:mnist_autocorr1}
\end{figure}

\begin{figure}[h]
\centering
   \includegraphics[width=0.3\linewidth]{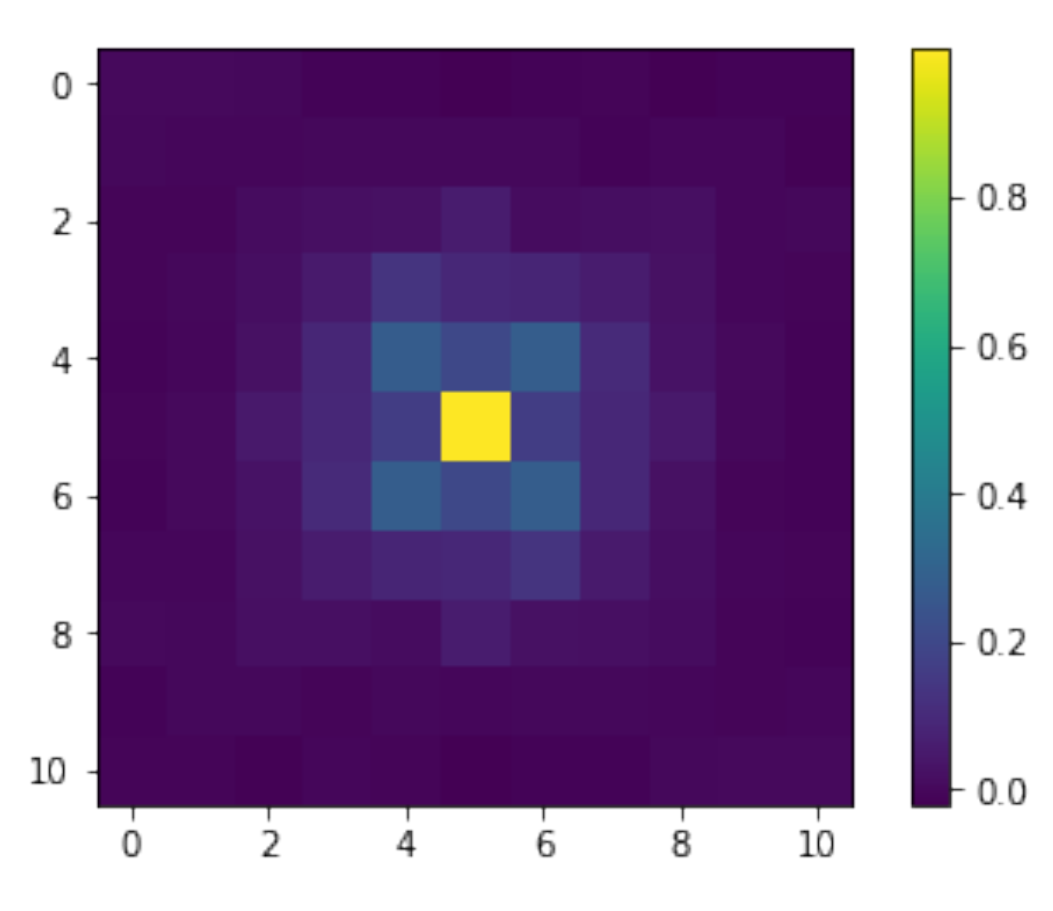}
\caption{Autocorrelation of the gradients: ImageNet, VGG16-bn, $11\times 11$ kernel}
\label{fig:imagenet_autocorr1}
\end{figure}

We provide further evidence for considering a non-identity covariance for modelling the gradient covariance. In figure \ref{fig:mnist_autocorr1}, we plot the autocorrelation of the true gradients from 100 images sampled from the MNIST test set for kernel size of $11\times 11$ for the LeNet model. In figure \ref{fig:imagenet_autocorr1}, we plot the autocorrelation of the third channel of the true gradients from 50 images sampled from the ImageNet validation set for kernel size of $11\times 11$ for the VGG16-bn model. The autocorrelation plots show the correlation across dimensions and across images to be substantial so as to provide even more evidence and reason for the gradient model we have considered in this paper.

\end{document}